\documentclass[letterpaper,twocolumn,fleqn]{article} 

\usepackage{ist}
\usepackage{comment}
\usepackage{graphicx}
\usepackage{amsmath}
\usepackage{amssymb}
\usepackage{nccmath} %%%%
\usepackage[ruled,vlined]{algorithm2e}
\title{SalyPath360: Saliency and Scanpath Prediction Framework for Omnidirectional Images}

\author{ Mohamed Amine KERKOURI\textsuperscript{1}, Marouane TLIBA\textsuperscript{1}, Aladine CHETOUANI\textsuperscript{1}, Mohamed SAYEH\textsuperscript{2} \\ \textsuperscript{1}Laboratoire PRISME, Université d'Orléans, Orléans, FRANCE \\ \textsuperscript{2}University of Oran 1 ORAN, ALGERIA}

\date{} % date has an empty field.

\begin{document} 

\maketitle 

\thispagestyle{empty} % prevents the first page to be numbered

%%%%%%%%%%%%%%%%%%%%%%%%%%%%%%%%%%
% Abstract
%%%%%%%%%%%%%%%%%%%%%%%%%%%%%%%%%%

\begin{abstract}
This paper introduces a new framework to predict visual attention of omnidirectional images. The key setup of our architecture is the simultaneous prediction of the saliency map and a corresponding scanpath for a given stimulus. The framework implements a fully encoder-decoder convolutional neural network augmented by an attention module to generate representative saliency maps. In addition, an auxiliary network is employed to generate probable viewport center fixation points through the $SoftArgMax$ function. The latter allows to derive fixation points from feature maps. To take advantage of the scanpath prediction, an adaptive joint probability distribution model is then applied to construct the final unbiased saliency map by leveraging the encoder decoder-based saliency map and the scanpath-based saliency heatmap. The proposed framework was evaluated in terms of saliency and scanpath prediction, and the results were compared to state-of-the-art methods on Salient360! dataset. The results showed the relevance of our framework and the benefits of such architecture for further omnidirectional visual attention prediction tasks. 
\end{abstract}

%%%%%%%%%%%%%%%%%%%%%%%%%%%%%%%%%%%%
% Overall Document Guidelines: Head
%%%%%%%%%%%%%%%%%%%%%%%%%%%%%%%%%%%%
\section{1. Introduction}
Virtual Reality (VR) applications provide high quality of immersive user experiences.  Most of VR applications are in the form of 360 video, whereas the frames are represented under a new format of multimedia content called omnidirectional image.
These images cover the whole spherical viewing space ($360^\circ\times180^\circ$), where the user has the freedom of attending to any direction just by pointing his head to any direction.
%These head movements could be represented using the Euler angles that correspond to viewer head rotation around the 3D space axes. 
The viewport of the $360^\circ$ image is defined by the device-specific viewing angle (typically 120 degrees), which delimits horizontally the scene from the head direction center, called viewport center.
The rendering of $360^\circ$ viewport of the images is supported by many types of sphere to plane coordinates mapping transformations, EquiRectangular Projection (ERP) is one of the most widely used formats of uniform quality mapping projection \cite{ERP-Adriano,uniform}. 
It projects the spherical content to a single high resolution 2D plane, where the longitudinal and latitudinal sphere coordinates are represented on the horizontal and vertical ERP axes, respectively.

Unlike the traditional fixed viewport delivery of 2D content, the immersive experience is delivered using recent technologies such as Head Mounted Display (HMD). They are empowered by the ability to investigate spherical space, enabling them to have the best realistic immersive experience with a high consumption of resources. Therefore, the capacity to predict the attended viewport that corresponds to the orientations of the head movements beforehand, helps to optimize the delivery process and to provide a higher Quality of Experience (QoE) to the viewer \cite{QoE}. 

This can be achieved through the prediction of the human visual attention that reflects the most interesting regions within the field of view of the users. This natural mechanism allows humans to explore complex scenes effortlessly and devotes their limited perceptual resources to the most pertinent subsets of received sensory information \cite{VisionAtt}. 
The attractive regions, often called salient regions, are usually represented in a heatmap (i.e saliency map). This map models the distribution of the gaze fixations describing the probability that a pixel is salient. 
The saliency maps are generated by processing the scanpaths of different viewers, which are defined as a sequence of successive fixation points of the viewer's gaze while exploring the image \cite{scanpathDef}. 

Unlike conventional 2D images, the users are exposed in omnidirectional images to a larger degree of freedom. Visual attention modeling in $360^\circ$ content is conducted by predicting the center of most probable attended viewports, which reflect the trajectory of the head movements. Moreover, we assume as in \cite{XuMSong} that human attention toward omnidirectional content is governed by some statistical bias, as they tend more to equatorial and frontal regions than others, referred as equator bias.

Studies on omnidirectional content were pioneered by the work of Bogdanova et al. \cite{Iva,Iva1} where the spherical static saliency map was generated by normalizing and merging chromatic, intensity and three cue conspicuities. They also created a motion spherical saliency map through a motion pyramid decomposition. At the last stage, the two resulting maps were fused to produce the spherical frame saliency map. %In \cite{BattistiF}, the authors combined both viewport relevant high level features and low-level features from hue, saturation and Graph-Based Visual Saliency (GBVS) \cite{GBVS} model. %In \cite{FangY}, the authors extracted low-level features as color, texture, luminance and boundary connectivity from the ERP images.
In \cite{DeAbreuA}, the authors introduced a Fused Saliency Map (FSM) to predict visual attention on ERP omnidirectional images by adopting the well-known 2D saliency model SALICON \cite{SALICON}. In \cite{GBVS360BMS360}, the authors extended both the 2D Boolean Map Saliency (BMS) \cite{BMS} and GBVS \cite{GBVS} by incorporating the characteristics of ERP images.

%Due to the availability of large scale datasets \cite{salient360!images,Iva}, various Deep Neural Network (DNN) based saliency prediction models for $360^\circ$ images have been introduced.
With the high performance of deep neural networks in imaging, various saliency prediction models were introduced.  
In \cite{Nguyen}, the authors fine-tuned a 2-Dimensional static saliency model, called PanoSalNet, on ERP frames for the task of head movement prediction. The resulting saliency map is further enhanced by prior statistical bias. 
In \cite{ChaoFung}, the authors fine-tuned also a 2D static model, called SalGAN, on $360^\circ$ image dataset using cube-map projection and a new objective function, leveraging the combination of three saliency measures. Instead of using a supervised  approach to learn saliency from labeled data.% in \cite{XuMSong} the authors adopted deep reinforcement learning to predict the position of head movements. 
In \cite{AtSal}, the authors proposed a novel attention-based architecture that adapts the encoded latent vector to the characteristics of omnidirectional images through the extended receptive field. They exploited a Cubic Map Projection (CMP) to improve prediction on polar regions. 
In \cite{2DSal3D}, the authors proposed a new framework that applies existing 2D saliency models on ERP images without requiring in-depth adaptation of the prediction algorithms. They adopt an adaptive weighted joint probability distribution on different kinds of projection of omnidirectional images. 

Scanpath prediction methods are more scarce in literature, even more so in 360$^\circ$ methods. In \cite{Heuristic360}, the authors proposed a model that uses low level features to produce a saliency map. The resulting map is then binarized and fixation points are generated from the obtained binary map using a clustering method. A graph is constructed from the fixation points. The scanpath generated by maximizing the sum of transfer probabilities on the graph edges. In \cite{saltinet}, the authors proposed a hybrid method, that uses deep neural networks and heuristic methods. They use an encoder-decoder network to generate a static saliency volume, from which scanpaths are generated through stochastic techniques. In \cite{pathgan}, the authors proposed a deep neural network, that uses an encoder-decoder network and Long Short Term Memory (LSTM) layers to generate scanpaths, combined with a adversarial training by Generative Adversarial Network (GAN) architecture. 

In this paper, we introduce a new framework to predict visual attention of omnidirectional images. The proposed architecture, called \textit{SalyPath360}, allows the simultaneous prediction of saliency map and a corresponding scanpath for a given stimulus. 

%The framework implements an encoder-decoder fully convolutional neural network, augmented by an attention module to generate representative saliency maps. In addition, an auxiliary network is employed to generate probable viewport center fixation points. We use a sequence of convolutional layers, followed by a $Soft-Argmax$ function to predict fixation points, producing an ordered sequence of Head-Movement positions. 

%To take advantage of the scanpath prediction, an adaptive joint probability distribution model is then applied to construct the final unbiased saliency map by leveraging the encoder decoder-based saliency map and the scanpath-based saliency heatmap. The experimental results on Salient360! dataset \cite{salient360!images}  show the relevance of our framework and the benefits of such architecture for further omnidirectional visual prediction tasks.% The code will be available at : \href{https://github.com/Papercode971/SalyPath360}{https://github.com/Papercode971/SalyPath360}.%\footnote{The github will be accessible after the review step in order to preserve the double blind review process.}}

The main contributions of this paper are as follows:
\begin{itemize}
    \item Presenting a Neural Network that predicts saliency map and scanpath for $360^\circ$ in simultaneous and parallel manner. 
    \item Unlike existing methods, scanpaths are predicted from  the refined internal features of a proven saliency prediction model.
    \item Improving the saliency prediction by using a joint probability mixture between the saliency map predicted by the network and saliency map constructed from the predicted scanpath. 
    \item Use the $SoftArgMax$ function to predict head scanpaths, and train the network seamlessly.
\end{itemize}

The rest of this paper is organised as follows: Section 2. provides a detailed description of the proposed approach, while Section 3 compares the performance of our approach against state-of-the-art methods. Finally, we conclude our study and discuss the possibility of future improvements in Section 4. 

\section{2. SalyPath360 Framework}
\label{ProposedMethod}

\begin{figure*}[ht]
\makebox[\linewidth]{
\includegraphics[width= \linewidth] {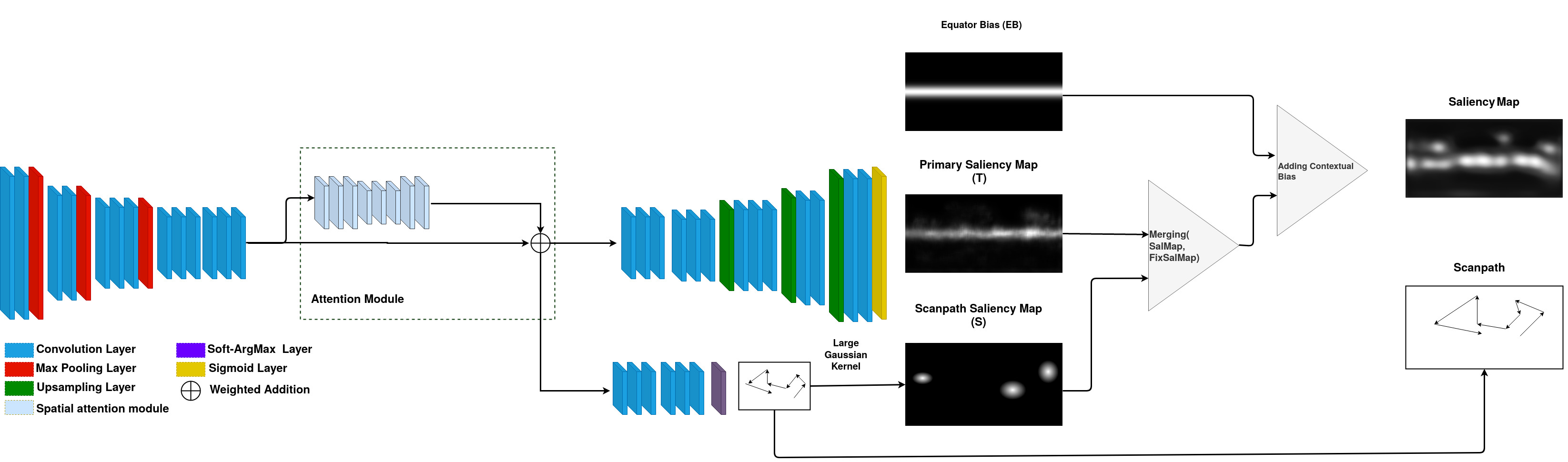}
}
\caption{Proposed framework called SalyPath360.} 
\label{fig:arch}
\vspace{-3mm}
\end{figure*}

In this section, we describe in detail the proposed framework (see Fig.~\ref{fig:arch}) which is mainly composed of an encoder-decoder network augmented by a spatial attention module and an auxiliary network that takes the intermediate features at the bottleneck of the encoder-decoder network to generate the corresponding scanpath. In addition, the primary saliency map predicted by the encoder-decoder architecture and the saliency map derived from the predicted scanpath are combined to generate a more representative saliency map. The architecture of each network as well as the attention module used and the merging process applied are described below. 
%
%The model a VGG-based encoder-decoder architecture with a an auxiliary branch for scanapath prediction and an attention module at the bottleneck.

\subsection{2.1. Encoder-Decoder Network for Saliency Prediction}

% Encoder - decoder model

%The attention module~\cite{} takes a $360^\circ$ image in an ERP representation format and predicates the first saliency map. This attention mechanism was used first in \cite{} and \cite{}, and attend to scan the different regions of the feature maps, in order to capture a global representation covering the integrity of the stimulus.
%The use of the spatial attention module enables to refine the extracted features from the encoder. Indeed, a succession of convolutional and max pooling layers broaden the receptive field and the Sigmoid activation followed by an upsampling layer generate a well  representative salient filter. 

The encoder-decoder network is used to predict the primary saliency map of $360^\circ$ images and generate high level intermediate features exploited by the auxiliary network. Inspired by \cite{AtSal} attention stream, the encoder and decoder are composed of four blocks of convolutional layers, interleaved respectively by a ($2\times2$) max-pooling layer and a ($2 \times 2$) up-sampling layer. At the bottleneck of our network, an attention module is employed to further refine the intermediate features used by the decoder and auxiliary networks. %This kind of attention mechanism has proven its usefulness for different tasks \cite{ChenAtt}, \cite{Xu2015}. 
This module takes the representational feature maps and predicts a single channel heatmap, which captures a global representation covering the integrity of the stimulus. It also expands the receptive field of the encoder from $(244\times244)$  to $(676\times676)$. This expansion is vital to cover the large $360^\circ$ images. The resulting heatmap is then used as a filter that refines the feature maps generated by the encoder as follows:% Eq.~\ref{eq:input_decode1}. %
%\vspace{-2mm}

\begin{equation}
  X^\prime = \gamma \otimes X \otimes Att_S( X) + X 
  \label{eq:input_decode2}
\end{equation}
where  \textit{$Att_S$} is the spatial attention module function. $X$ is the input feature maps given by the encoder. $X^\prime$ is the refined feature maps passed to the decoder and auxiliary networks. $\gamma$ is a learnable parameter, and $\otimes$ represents the element-wise multiplication. 

%It is worth noting that other deep learning-based saliency models can be used. However, models that use an encoder-decoder architecture facilitates the localization of our auxiliary network (i.e bottleneck of the model).

% Indeed, a succession of convolutional and max pooling layers broaden the receptive field and the Sigmoid activation followed by an upsampling layer generate a well globally representative heatmap \cite{REF}.
%The module broadens the field of view of the encoder and covers a receptive field of size  $676 \times 676$ pixels. 

%Thus, the encoder network boosted by the attention module extracts high level features for the stimulus, and passed them to an auxiliary network for the scanpath prediction and specific second saliency map generation. 

\subsection{2.2. Auxiliary Network for Scanpath Prediction}
% decoder 

An auxiliary network is also used to generate a scanpath for a given stimulus by mainly leveraging the encoding ability of the encoder-decoder network. %Table \ref{tab:aux_net} gives an overview of this network. 
It consists of $3 \times 3$ convolutional layers, each activated by a $ReLU$ function. The last layer is composed of $100$ features maps (i.e. 100 heatmaps), set in accordance to the number of fixation points per user of the considered dataset \cite{salient360!images} (see Section \ref{DATASET}).
$Soft-Argmax$ (SAM) \cite{SAM} is then used to estimate the coordinates of a fixation points from each feature map as follows :% using the highest activation are then extracted from the heatmaps using the $Soft-Argmax$ function \cite{SAM}, defined as follows: 

%\vspace{-1mm}
\useshortskip
\begin{equation}
  SAM(x) =  \sum_{i=0}^{W} \sum_{j=0}^{H} \frac{e^{\beta x_{i,j}}}{\sum_{i^\prime =0 }^{W} \sum_{j^\prime = 0 }^{H} e^{\beta x_{i^\prime ,j^\prime }}}({\frac{i}{W},\frac{j}{H}})^T
  \label{eq:SAM}
\end{equation}
where ${i,j,i^\prime ,j^\prime}$ iterate over pixel coordinates. ${H,W}$ represent the height and width of the feature map, respectively. $x$ is the input feature map and $\beta$ is a parameter adjusting the distribution of the softmax output.

%Each predicted coordinate represents a fixation point of the scanpath,  each one extracted from a given feature map. 
As this SAM function is differentiable \cite{SAM}, it permits our model to be trained seamlessly unlike the discrete $Argmax$ function. It also allows a sub-pixel accuracy, avoiding thus the use of upsampling layers to increase the size of the feature maps, and saving resources.

\begin{comment}

\begin{table}[htbp]
\small
\begin{center}
\begin{tabular}{ c c c   }
\hline
\textbf{Layer} & \textbf{Output shape} & \textbf{$\#$parameters }     \\ 
\hline
 Conv2D\_1 +ReLU & 512x40x20 &  2,359,808    \\ 
 \hline
 Conv2D\_2 +ReLU & 512x40x20 &  2,359,808    \\ 
 \hline
 Conv2D\_3 +ReLU & 512x40x20 &  2,359,808    \\ 
 \hline
 
 Conv2D\_4 +ReLU & 256x40x20 & 1,179,904    \\ 
 \hline
 
 Conv2D\_5 +ReLU & 256x40x20 &  590,080    \\ 
 \hline
 Conv2D\_6 +ReLU & 100x40x20 &  230,500    \\ 
 \hline
 Conv2D\_7 +ReLU & 100x40x20 &  90,100    \\ 
 \hline
 
 SoftArgMax & 100x2  &  0     \\ 

 \hline\\
\end{tabular}
\caption{\label{tab:aux_net}Auxiliary network architecture.}
\end{center}
%\vspace{-5mm}
\end{table}
\end{comment}

\subsection{2.3. Merging Probability Distribution of Unbiased Saliency Map Prediction}
\label{sec:JPD}

Let us consider the predicted saliency map $T$ constructed from encoder-decoder architecture (i.e. primary saliency map), and the saliency map $S$ generated from the predicted fixation points by an adequate Gaussian kernel.
%Inspired by \cite{Winkler}, Where authors defined a joining distribution model that merges independent signals. 
We notice that $T$ and $S$ could be combined together to get a joint probability distribution~\cite{Cooke}, making thus a final representative saliency map straightened by the most probable viewport center positions following : 

\begin{comment}

%the adopted weighted combination for $n$ distributions $\{p_1,\cdots, p_n\}$ is considered as a joint probability $J_r$ inspired from the merged term~\cite{2DSal3D} introduced to underlay the contraction of a well representative saliency distribution. Inspired by these studies, the joint probability distribution $J(T,S)$ for the saliency maps $T$ and $S$ as follows:

%\vspace{-1mm}
%

\begin{equation}

    J_r = \max\limits_p \times \bigg( {\sum_{i=1}^{n} \alpha_i{\bigg({\frac{p_i}{\max\limits_p}}\bigg)^r}} \bigg)^{1/r} \enspace.
    \label{eq:r-norm-probability-formula}
\end{equation}
%where $r$ is the r-norm for the weighted mean, the terms $\{\alpha_1,\cdots, \alpha_n\}$ are the weight coefficients with $\sum_{i=1}^{n} {\alpha_i} = 1$, and $\max\limits_p = max(|p_i|,\cdots,|p_n|)$ is the factorization and normalization formula.
\end{comment}
%\useshortskip

\begin{equation}
%\scriptsize    
    J(T,S) = \max \limits_{_{_{T,S}}} \times \bigg( {{{\alpha \bigg(\frac{T}{\max\limits_{_{T,S}}}\bigg)^{k}+(1-\alpha)\bigg(\frac{S}{\max\limits_{_{T,S}}}\bigg)^{k}}}} \bigg)^{1/k} \enspace
    \label{eq:j1-probability-formula}
\end{equation}
 
Where $\max\limits_{_{_{T,S}}}$ is the maximum between the pair $(T,S)$. $k$ is the k-norm for the weighted mean formula and $\alpha$ represents the weight coefficients used to combine $T$ and $S$ distributions. The parameter $\alpha$  was set up throughout an experimental search and was set to $0.7$. 

For well-tuned $\alpha$ value, the joint probability distribution $J(T,S)$ models a saliency map that integrates  spatial and contextual features via the primary saliency map (T) on one side, and the scanpath saliency map (S) generated from the predicted head fixation points on the other. Nevertheless, the variation of head-based movement is about $180^\circ$ on the $Z$ axis, when considering a spherical referential $(X, Y, Z)$. But practice, the head variation has a limited range around the equator. Therefore, the computed joint probability distribution $J(T,S)$ should be corrected with another distribution that represents this phenomenon, called the equator bias $E$. 
%It is originally constructed as a column-wise symmetric bell curve distribution, where most of the probabilities fall on the equator region. 
The equator bias $E$ incorporates most probable head movement bias distribution. By regards with the experiments, we adopted the mean between the previous calculated join probability distribution and the new adjusted one with the equator bias $E$. %These predefined weights could be further fine-tuned using learning optimization. 
%As another motivation for incorporating the equator bias context, attentional saliency stream is first constructed for the head and eye movement tasks. 
The unbiased formula in our context is defined as follows:
\useshortskip
%\vspace{-1mm}
\begin{equation}
%\scriptsize    
    J^*(T,S) = \bigg(\frac{1}{2}\times J(T,S) + \frac{1}{2}\times J(T,S) \times L_s(E))\bigg)
    \label{eq:j-probability-unbiased-formula}
    %\vspace{-6mm}
    \useshortskip
\end{equation}
where $L_s(E)$ represents the pixel-wise linear-scaling of the equator bias $E$.

The applied merging process algorithm for the set of pixels ${T_{pixels}}$ can be summarized in Algo. \ref{algo:JoinProb}.
%of our architecture is constructed below for a given hyper-parameters values $\alpha$ and $k$, having as input two saliency $T$ and $S$ maps with a bias $E$ and gives as result an unbiased combined saliency map.   
%\vspace{-3mm}
%%
%\begin{comment}

\begin{algorithm}[h]
\SetAlgoLined
\KwResult{Joint probability distribution $J^*(T,S)$ }
$T$ Primary saliency map.\\
$S$ Scanpath-based heatmap.\\
$E$ Equator bias map.\\ 
 
 Compute the normalized linear scaling of bias $L_s$(E) \\
 L =  ${T_{pixels}}$ \\
 \For {$(l \in L)$} {
  $J(T,S)[l] \gets
\max\limits_{_{_{T,S}}} \times \bigg( {{{\alpha \bigg(\frac{T(l)}{\max\limits_{_{T,S}}}\bigg)^{k}+(1-\alpha)\bigg(\frac{S(l)}{\max\limits_{_{T,S}}}\bigg)^{k}}}} \bigg)^{1/k}$\;

$J^*(T,S)[l] \gets \bigg(\frac{1}{2}\times J(T,S)[l] + \frac{1}{2}\times J(T,S)[l] \times L_s(E)[l])\bigg)$ \;

}
\Return $J^*(T,S)$;

 \caption{\texttt{JSalyPath} (IN $T$,$S$,$E$ OUT $J^*(T,S)$)}
\label{algo:JoinProb}

\end{algorithm}
%\vspace{-5mm}

%\end{comment}
%\vspace{-6mm}
%the scanpath prediction through the refinement of the intermediate representational space. The latter representation was here optimized by the complex saliency loss function (see eq. \ref{SalL}). 

%Therefore, the complex loss function used for the latter (see eq. \ref{SalL}) allows us to use a simple loss function for scanpath prediction.

%The Soft-ArgMax function allows a sub-pixel
%Because use of the Soft-ArgMax function allows a sub-pixel prediction on a continuous space, and the scanpath is spatially characterized mainly by the locations of it's fixations.Scanpath prediction is problem a regression one, for that end we used solely MSE loss, as described in Eq.\ref{eq:SPL}.         
%as the scanapath is characterized by locations of it's fixations. and their prediction is classic regression problem,  that is why we used the MSE function. 
\begin{table*}[ht!]
\small
\begin{center}
\begin{tabular}{ c  c  c  c  c  c  c  c c}
\hline
\textbf{Model} & \textbf{Auc Judd $\uparrow$} & \textbf{Auc Borji$\uparrow$} & \textbf{NSS$\uparrow$}& \textbf{CC$\uparrow$}& \textbf{SIM$\uparrow$} & \textbf{KLD $\downarrow$}  \\ 
\hline

ATSal\cite{AtSal} &0.8479&0.8121&1.7516&0.6214&0.5748&1.1571\\
\hline

 Two-stream \cite{twostream} &  0.7931 & 0.7564 & 1.6249 & 0.5857 & 0.5857 & 0.8585 \\
 \hline

SalyPath360 (Our method)  & \textbf{0.8610} & \textbf{0.8199} & \textbf{1.8552} & \textbf{0.7194} & \textbf{0.6383}    &  \textbf{0.8405}  \\ 
\hline \\
\end{tabular}
\caption{\label{tab:sal_metrics} Table 1. Saliency prediction Comparison. Best results are highlighted in bold.}
\end{center}
\vspace{-9mm}
\end{table*}

%final, AUC_Judd, 0.8501907424624433
%final, AUC_Borji, 0.8143054316164474
%final, NSS, 1.7547685904118888
%final, CC, 0.6278294049073337
%final, SIM, 0.5791680333808904
%final, KLD, 1.1105727952394693

\subsection{2.4. Training}

The encoder-decoder and the auxiliary networks were trained through different loss functions. More precisely, the encoder-decoder network based on ATSal \cite{AtSal}, which  was trained on the Head-Eye movement datasets, was fine-tuned for our task of head movement prediction using the following loss function $L_1$: 
\useshortskip
\begin{equation}
\begin{split}
    \textit{$L_1$} = 0.8 \times  KLdiv(y,\hat{y}) + 0.2 \times  BCE(y,\hat{y} ) \\ - 0.2 \times NSS(y_{fix},\hat{y})
     \label{eq:SAlL}
\end{split}
\end{equation}
where $KLdiv$ is the Kullback-Leibler Divergence, $BCE$ is the Binary Cross Entropy and $NSS$ is the Normalized Scanpath Saliency. $y$ and $\hat{y}$ represent the ground truth saliency map and the predicted saliency map respectively, while $y_{fix}$ is the ground truth fixation map.

Each term of this loss function was chosen for its own influence on the convergence of the network. Indeed, $KLD$ and $BCE$ functions minimize the distance between the distributions of the output and ground truth, while $NSS$ is a similarity metric for saliency and it is used as bias term, allowing the model to seize the saliency specific representations. 

As the auxiliary branch aims to predict the coordinates of fixation points, the task can be seen as a regression problem. In addition, this branch relies upon the feature representations given by the encoder-decoder which is trained by the more complex loss function described-above. As such, this loss (i.e. $L_1$, see Eq. \ref{eq:SAlL}), has an indirect impact on the auxiliary branch during the training step. Therefore, we chose the Mean Squared Error ($MSE$) as a simple loss function $L_2$ to train this branch, defined  as follows:
%\vspace{-1mm}
\useshortskip
\begin{equation}
    L_2 = \frac{1}{N} \sum_i (p - \hat{p} )^2
    \label{eq:SPL}
\end{equation}
where $p$ is the ground truth scanpath and $\hat{p}$ is the predicted scanpath, while $N$ is the number of fixation points of the corresponding scanpaths.

The 2 branches were trained consecutively by first fine-tuning the encoder-decoder network as it has a great influence on the accuracy of the auxiliary network. Then, the auxiliary network is trained from scratch while freezing the weights of the encoder in order to not affect the saliency prediction.

%\vspace{-3mm}
\section{3. Experiments}
\label{Experiments}

In this section, we evaluate the ability of our model to predict saliency maps and scanpaths. We first describe the used dataset. Then, qualitative and quantitative results are presented and compared with state-of-the-art methods.

\subsection{3.1. Dataset}
\label{DATASET}

\textbf{Salient360!} \cite{salient360!images} dataset is one of the most used datasets for predicting saliency of omnidirectional images. It was proposed as a part of the 2018 Salient360! Challenge and it is composed of 85 omnidirectional images with their corresponding saliency maps and scanpaths. There are about 36 scanpaths per image where each scanpath has 100 fixation points represented by their coordinates and timestamp. 

The dataset was split into training-testing sets without any overlap according to the same protocol use in \cite{AtSal}. The training set is composed of 70 images, while the test set is composed of 15 images, representing $82\%$ and $18\%$ of the dataset respectively. This choice of split was done in accordance with the common practices used in other papers \cite{AtSal} using this dataset. It is worth noting that for a fair comparison, all the compared models were evaluated using the same partition. To the best of our knowledge, Salient360! is the only publicly available dataset that provides the scanpaths in addition to the saliency maps for each $360^\circ$ image. 

\subsection{3.2. Saliency Prediction}

\begin{figure*}[h!]
\makebox[\linewidth]{
\includegraphics[width= \textwidth]{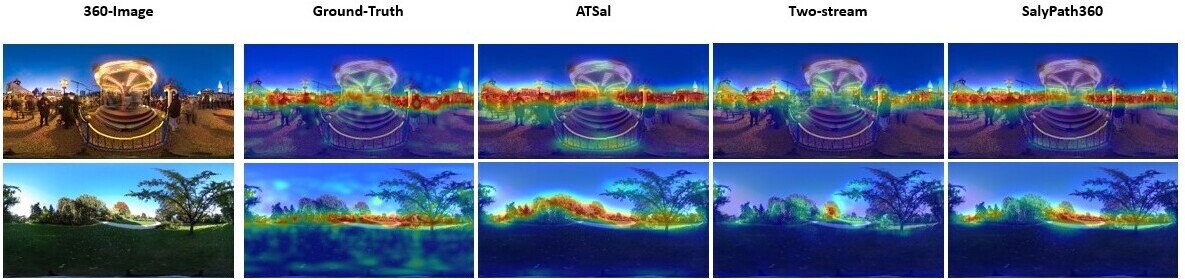}
}
\caption{Qualitative comparison of saliency maps predicted through the proposed SalyPath360 model with state-of-the-art models.} 

\label{fig:Q1}
\end{figure*}

\begin{figure*}[h!]
\makebox[\linewidth]{
\includegraphics[width=\textwidth]{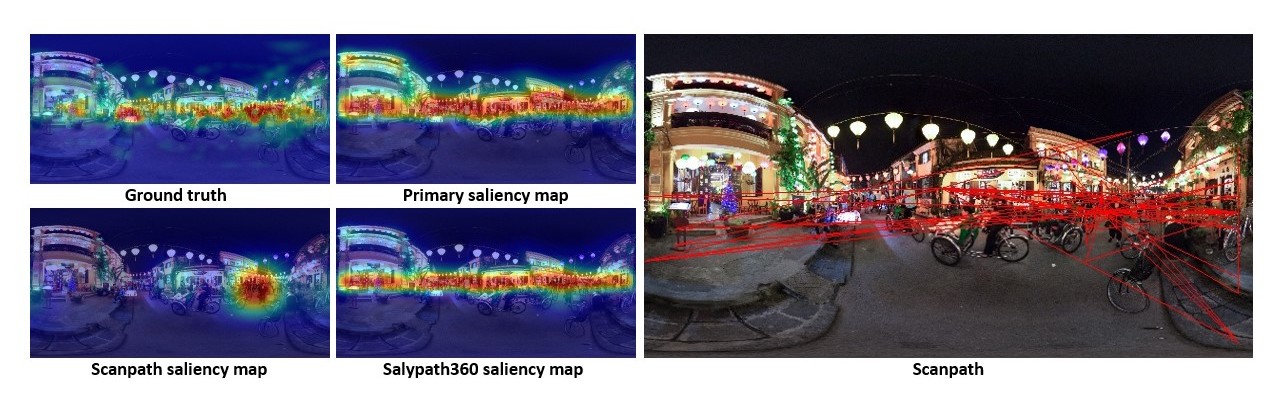}
}
\caption{Qualitative results: On the left part, we show the ground truth saliency map as well as those obtained by the encoder-decoder network, the auxiliary network and the merged map. On the right part, we show the corresponding predicted scanpath using SalyPath360.} 

\label{fig:Q2}
%\vspace{-6mm}
\end{figure*}

\begin{comment}
\begin{figure*}[h!]

\includegraphics[width=\textwidth]{SP_viz.jpg} \\

\caption{Qualitative results: Results and comparison of scanpaths obtained by SalyPath360 and other models.}

\end{figure*}
\end{comment}

To evaluate the saliency prediction effectiveness of our method, we employed commonly used saliency metrics: $Auc\_Judd$, $Auc\_Borji$, $NSS$ , $CC$, $SIM$, $KLD $ \cite{Metrics}. The results are compared to state-of-the-art saliency models: Two-Stream \cite{twostream} that achieved the best result on Salient360 challenge \cite{salient360challenge} and ATSal \cite{AtSal} that reached high prediction results  Salient360 dataset. It is worth noting that for \cite{twostream}, we used the model published on the lead board of the Salient360! Challenge as Wuhan University. For \cite{AtSal}, we used results provided by the authors for their still image model to calculate results. 

%It is worth noting that for \cite{twostream}, we used the version of code used for still images competition published in  the leader board as Wuhan university \footnote{https://github.com/zhangkao/IIP_Salient360_2018}. As for \cite{AtSal} we used results provided by the authors for still images version to calculate the metrics results..
Table \ref{tab:sal_metrics} shows the results obtained. Best results are highlighted in bold. As can be seen, the proposed model outperforms all the compared saliency methods, including ATSal. For $Auc\_Borji$, we achieve a slight improvement over the state-of-the-art results, while for the other metrics high improvements are noted with a considerable improvement for $NSS$ and $CC$. Fig. \ref{fig:Q1} shows a qualitative comparison between saliency maps predicted using our framework and state-of-the-art models as well the ground truth saliency maps. As can be seen, the saliency maps generated by our framework are closer to the ground-truth.   

\subsection{3.3. Scanpath Prediction}

In this section, we evaluate the results obtained by our framework, regarding both the scanpath and the final saliency prediction using common metrics. More precisely, we employed a vector-based metric called $Jarodzka$ \cite{MM} which compares the similarity between the scanpaths and the hybrid $NSS$ \cite{NSS} metric, which compares the scanpath with the ground truth saliency map. For the former, we applied the code used during the Salient360! challenge \cite{salient360challenge}, while disregarding the temporal element which is not predicted by our framework. We also compare the performance of our framework with state-of-the-art models: PathGan \cite{pathgan} and SaltiNet \cite{saltinet}.

\begin{table}[htbp]
\small
\begin{center}
\begin{tabular}{ c c c  }
\hline
\textbf{Model} & \textbf{Jarodzka $\downarrow$ } & \textbf{NSS $\uparrow$ }   \\ 
\hline
 PathGan\cite{pathgan} &   0.1777 & -0.1518     \\ 
 \hline
 SaltiNet\cite{saltinet} & 0.2621 & 0.0834        \\
 \hline
 SalyPath360 (Our method) & \textbf{0.1363}  &  \textbf{0.2896}      \\ 
 \hline\\
\end{tabular}
\caption{\label{tab:scanpath_results}Table 2. Scanpath prediction comparison. Best results are highlighted in bold.}
\end{center}
\vspace{-6mm}
\end{table}

Table \ref{tab:scanpath_results} presents the results obtained with the best results are highlighted in bold. As can be seen, our model achieves the best results on $Jarodzka$ and $NSS$. PathGan obtains the second best results on $Jarodzka$, while SaltiNet achieves better results on $NSS$ metric, outperforming PathGan.

%{\color{blue} We also analyse the statistical significance of difference through an ANOVA test. Fig. \ref{fig:Anova} shows the boxplots obtained ... As can be seen, the distributions are quite different with difference ... These differences are statically significant since the p-values obtained are lower than 0.05.}

\begin{figure}[ht]
\makebox[\linewidth]{
\includegraphics[width= \linewidth] {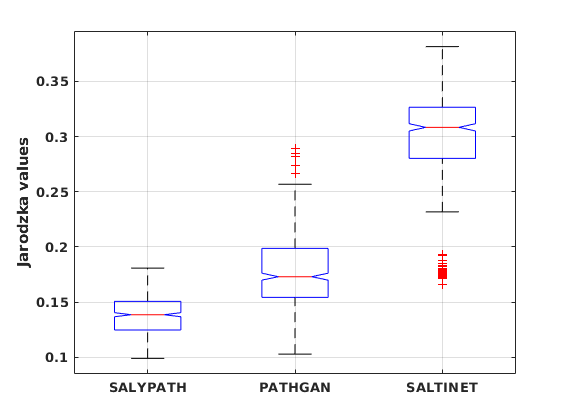}
}
\caption{Results of the one-way Anova test for the $Jarodzka$ values obtained for each compared method.} 
\label{fig:Anova}
\end{figure}

A one way Analysis of Variance (ANOVA) test between groups is also applied to show if the differences between the distributions of the $Jarodzka$ values obtained for each compared method are statistically significant. 
Fig. \ref{fig:Anova} shows the corresponding boxplots where the middle red mark represents the median value, and the contours of the box are the $25^{th}$ and $75^{th}$ percentiles. 
The extremities of the whiskers correspond to the minimum and maximum values without considering the outliers. 
The outliers are represented by red crosses and correspond to data points which are further than two or three standard deviations. 
As can be seen, the distributions are quite different with the smaller median value and data scatter (i.e. standard deviation), followed by those of PathGan. We then computed the p-value between the proposed method and each of the compared method (i.e. SaltiNet and PathGan). The p-values were lower than the significance level (i.e. 0.05) in both cases, indicating that the differences between the distributions of the compared method are statistically significant. %The results for our model are shown be less scattered compared the  the other models, and with no outliers.  
%... As can be seen, the distributions are quite different with difference ... These differences are statically significant since the p-values obtained are lower than 0.05.
Fig.\ref{fig:Q2} shows qualitative comparison of the predicted and ground-truth saliency maps as well as those obtained by the encoder-decoder and the auxiliary networks. We also show the corresponding predicted scanpath using SalyPath360. As can be seen, the scanpath predicted through the proposed auxiliary network spans most of the salient regions, while maintaining the bias of equator and frontal regions, but does not visually qualify to be probable. This disparity  between the quantitative results obtained through $Jarodzka$ metric and the qualitative results shows the limitations of the metric concerning the comparison.

\subsection{3.4. Impact of the joint probability merging}

\begin{table*}[ht!]
\small
\begin{center}
\begin{tabular}{ c  c  c  c  c  c  c  c c}
\hline
\textbf{Model} & \textbf{Auc Judd $\uparrow$} & \textbf{Auc Borji$\uparrow$} & \textbf{NSS$\uparrow$}& \textbf{CC$\uparrow$}& \textbf{SIM$\uparrow$} & \textbf{KLD $\downarrow$}  \\ 
\hline

Scanpath Saliency map (S) & 0.7746 & 0.7400 & 1.4706  & 0.3629 &	0.4473 &	2.0983  \\
\hline
Merging map J(T,S) Eq. \ref{eq:j1-probability-formula} & 0.8501 & 0.8143 & 1.7547 & 0.6278 & 0.5791 & 1.1105 \\
\hline
SalyPath360 (final map)   & 0.8610 & 0.8199 & 1.8552 & 0.7194 & 0.6383 &  0.8405  \\ 
\hline \\
\end{tabular}
\caption{\label{tab:impact} Table 3. Impact of joint probability module. }
\end{center}
\vspace{-10mm}
\end{table*}

To assess the efficiency of the different components of our framework, we evaluated the saliency maps obtained from the predicted scanpaths, and the saliency maps obtained after the merging without the Equator Bias.  

Table \ref{tab:impact} displays the results obtained during this study.
The maps generated from the scanpath (S) show poor performance for metrics comparing the distributions (i.e. KLD, SIM, CC) while the results for the location-based metrics  \cite{metrics_mean} (i.e. NSS, Auc\_Judd, Auc\_Borji) show results closer to the comparing models. It is worth noting that the map (S) represents the saliency of a single scanpath. Therefore, the results obtained are satisfactory compared to the ground truth maps aggregating 32 scanpaths. After the merging process (i.e. predicted scanpath generated saliency map and the predicted saliency map), we evaluated the results obtained for J(T,S) in Eq.\ref{eq:j1-probability-formula}. The results on all the metrics showed a slight improvement compared to ATSal and Two-Stream models. While the results after using the Equator Bias show a significant improvement to those after the merging, indicating the beneficial effects of the using scanpath prediction and the merging module.%, and greatly emphasising the importance of the equator bias.      

% scanpath maps  --> results for the NSS  , Auc_Judd  are good but still low, results for other metrics  are not as good     
% Merging --> the resulst for all the metrics are better then those for Two-Stream and ATSal.  

% conclusion of the study 
% 
%  The use of saliency maps  from  scanpaths  improves the results 
%  The use of EB improves the model further 

\section{4. Conclusion}
\label{Conclusion}
In this paper, we introduced a new framework that simultaneously predicts saliency and scanpath for omnidirectional images. The proposed model is composed of an encoder-decoder convolutional neural network for saliency prediction strengthened by an attention module, and an auxiliary network to predict the corresponding scanpath. The latter is then convolved by a Gaussian filter to derive a heatmap. A merging adaptive probability model was finally added at the end of the framework to extract a representative saliency map through the predicted saliency map and the generated scanpath-based heatmap as well as an equator bias map. In our experiments, we trained and tested the framework on the Salient360! dataset. The results were compared with state-of-the-art saliency and scanpath models, showing the effectiveness of the proposed framework for both tasks. The qualitative results also confirmed the efficiency of our model. 

As perspectives, we will further improve the framework by incorporating the temporal dimension, and try to consider the inter-dependence between successive fixation points.

%\section{Acknowledgments} 
%add the acknowledgement section here

% To start a new column (but not a new page) and help balance the last-page
% column length use \vfill\pagebreak.

%%%%%%%%%%%%%%%%%%%%%%%%%%%%%%%%%%
% Bibliography
%%%%%%%%%%%%%%%%%%%%%%%%%%%%%%%%%%

{\small
\bibliographystyle{ieee_fullname}
\bibliography{biblio}
}

\begin{comment}

\end{comment}

%%%%%%%%%%%%%%%%%%%%%%%%%%%%%%%%%%
% Biography
%%%%%%%%%%%%%%%%%%%%%%%%%%%%%%%%%%
\begin{comment}

\begin{biography}
Please submit a brief biographical sketch of no more than 75 words. 
Include relevant professional and educational information as shown 
in the example below.

Jane Doe received her BS in physics from the University of Nevada (1977) 
and her PhD in applied physics from Columbia University (1983). Since 
then she has worked in the Research and Technology Division at Xerox 
in Webster, NY. Her work has focused on the development of toner adhesion 
and transport issues. She is on the Board of  IS\&T and a member of APS 
and SPIE.
\end{biography}
\end{comment}

\end{document}